\documentclass{article}

\usepackage{graphicx}
\usepackage{cite}

\usepackage{color}
\usepackage{amsmath}
\usepackage{epsfig}
\usepackage{url}

\newtheorem{thm}{Theorem}
\newtheorem{lem}{Lemma}

\newcommand{\qed}{\fbox{\rule[1pt]{0pt}{0pt}}}
\newcommand{\bsquare}{\hbox{\rule{6pt}{6pt}}}

\def\bfe{\mbox{\boldmath $e$}}

\def\bfs{\mbox{\boldmath $s$}}

\def\bfw{\mbox{\boldmath $w$}}
\def\bfx{\mbox{\boldmath $x$}}

\newcommand{\A}{{\bf A}}
\newcommand{\B}{{\bf B}}

\newcommand{\W}{{\bf W}}
\newcommand{\I}{{\bf I}}

\title{Finding Exogenous Variables in Data with Many More Variables than Observations}
\author{Shohei Shimizu\thanks{The Institute of Scientific and Industrial Research (ISIR), Osaka University,
Mihogaoka 8-1, Ibaraki, Osaka 567-0047, Japan. Email: sshimizu@ar.sanken.osaka-u.ac.jp}, Takashi Washio\thanks{Osaka University, Japan}, Aapo Hyv\"arinen\thanks{University of Helsinki, Finland} and Seiya Imoto\thanks{University of Tokyo, Japan}}
\date{}

\begin{document}

\maketitle

\begin{abstract}
Many statistical methods have been proposed to estimate causal models in classical situations with fewer variables than observations ($p$$<$$n$, $p$: the number of variables and $n$: the number of observations). However,  modern datasets including gene expression data need high-dimensional causal modeling in challenging situations with orders of magnitude more variables than observations ($p$$\gg$$n$). 
In this paper, we propose a method to find exogenous variables in a linear non-Gaussian causal model, which requires much smaller sample sizes than conventional methods and works even when $p$$\gg$$n$. 
The key idea is to identify which variables are exogenous based on non-Gaussianity instead of estimating the entire structure of the model. 
Exogenous variables work as triggers  that activate a causal chain in the model, and their identification leads to more efficient experimental designs and better understanding of the causal mechanism.
We present experiments with artificial data and real-world gene expression data  to evaluate the method.  
\end{abstract}

\section{Introduction}\label{sec:intro}
Many empirical sciences aim to discover and understand causal mechanisms underlying their objective systems such as natural phenomena and human social behavior. 
An effective way to study causal relationships is to conduct a controlled experiment. 
However, performing controlled experiments is often ethically impossible or too expensive in many fields including bioinformatics \cite{Bernardo05Nature} and neuroinformatics \cite{Londei06CP}. 
Thus, it is necessary and important to develop methods for causal inference based on the data that do not come from such controlled experiments. 

Many methods have been proposed to estimate causal models in classical situations with fewer variables than observations ($p$$<$$n$, $p$: the number of variables and $n$: the number of observations). 
A linear acyclic model that is a special case of Bayesian networks is typically used to analyze causal effects between continuous variables \cite{Pearl00book,Spirtes93book}. 
Estimation of the model commonly uses covariance structure of data alone and in most cases cannot identify the full structure (edge directions and connection strengths) of the model with no prior knowledge on the structure \cite{Pearl00book,Spirtes93book}.
In \cite{Shimizu06JMLR}, the authors proposed a non-Gaussian variant of Bayesian networks called LiNGAM and showed that the full structure of a linear acyclic model is identifiable based on non-Gaussianity without pre-specifying any edge directions between the variables, which is a significant advantage over the conventional methods \cite{Spirtes93book,Pearl00book}.  

However, most works in statistical causal inference including Bayesian networks have discussed classical situations with fewer variables than observations ($p$$<$$n$). 
Modern datasets including gene expression data need high-dimensional causal modeling in challenging situations with orders of magnitude more variables than observations ($p$$\gg$$n$)\cite{Bernardo05Nature,Londei06CP}. 
Here we consider situations in which $p$ is on the order of 1,000 or more, while $n$ is around 50 to 100. 
For such high-dimensional data, the previous methods are often computationally intractable or statistically unreliable. 

In this paper, we propose a method to find exogenous variables in a linear non-Gaussian causal model, which requires much smaller sample sizes than conventional methods and works even when $p$$\gg$$n$. 
The key idea is to identify which variables are exogenous instead of estimating the entire structure of the model. 
The simpler task of finding exogenous variables than that of the entire model structure would require fewer observations to work reliably. 
The new method is closely related to a fairly recent statistical technique called independent component analysis (ICA).

Exogenous variables work as triggers that activate a causal chain in the model, and their identification leads to more efficient experimental designs of practical interventions and better understanding of the causal mechanism. 
A promising application of Bayesian networks for gene expression data is detection of drug-target genes \cite{Bernardo05Nature}. 
The new method proposed in this paper can be used to find which genes a drug first affects and how it triggers the gene network.

The paper is structured as follows. 
We first review ICA and linear causal models in Section \ref{sec:principle}.
We then define a non-Gaussian causal model and propose a new algorithm to find exogenous variables in Section \ref{sec:method}. 
The performance of the algorithm is evaluated by experiments on artificial data and real-world gene expression data in Sections \ref{sec:exp} and \ref{sec:genedata}. Section \ref{sec:conc} concludes the paper. 


\section{Background principles}\label{sec:principle}

\subsection{Independent component analysis}\label{sec:ica}
Independent component analysis (ICA) \cite{Hyva01book} is a statistical technique originally developed in signal processing.
ICA model for a $p$-dimensional observed continuous random vector $\bfx$ is defined as
\begin{eqnarray}
\bfx &=& \A \bfs, \label{eq:ica}
\end{eqnarray}
 where $\bfs$ is a $p$-dimensional continuous random vector whose components $s_i$ are independent and non-Gaussian and are called independent components, 
and  $\A$ is a constant $p$$\times$$p$ invertible matrix. 
Without loss of generality, we assume $s_i$ to be of zero mean and unit variance. 
Let $\widetilde{\W}$$=$$\A^{-1}$. Then we have $\bfs$$=$$\widetilde{\W}$$\bfx$. 
It is known that the matrix $\widetilde{\W}$ are identifiable up to permutation of the rows \cite{Comon94}. 

Let $\widehat{\bfs}$$=$$\W\bfx$. 
A major estimation principle for $\widetilde{\W}$ is to find such $\W$ that maximizes the sum of non-Gaussianity of estimated independent components $\widehat{s}_i$, which is known to be equivalent to maximize independence between the estimates  when the estimates are constrained to be uncorrelated \cite{Hyva01book}. 
In \cite{Hyva99TNN}, the author proposed a class of non-Gaussianity measures:  
\begin{eqnarray}
J(\widehat{s}_i) &=& J_G(\bfw_i) = [ E\{G(\bfw_i^T\bfx)\} -E\{G(z)\}]^2, \label{eq:approxJ}
\end{eqnarray}
where $\bfw_i^T$ is the $i$-th row of $\W$ and is constrained so that $E(\widehat{s}_i^2)$$=$$E\{(\bfw_i^T\bfx)^2\}$$=$1 due to the aforementioned assumption on unit variance of $s_i$, $G$ is a nonlinear and non-quadratic function and $z$ is a Gaussian variable with zero mean and unit variance. 
In practice, the expectations in Eq. (\ref{eq:approxJ}) are replaced by their sample means. 
In the rest of the paper, {\it we say that a variable $u$ is more non-Gaussian than a variable $v$ if $J(u)$$>$$J(v)$}. 
The author of \cite{Hyva99TNN} further proposed an estimation method based on maximization of non-Gaussianity and proved a theorem to show its (local) consistency:
\begin{thm}\label{thm:Aapo}
Assume that the input data $\bfx$ follows the ICA model in Eq. (\ref{eq:ica}). 
Assume that $G$ is a sufficiently smooth even function. 
Then the set of local maxima of $J_G(\bfw)$ under the constraint $E\{(\bfw^T\bfx)^2\}$$=$1 includes the rows of $\widetilde{\W}$ for which the corresponding independent components $s_i$ satisfy the following condition $E\{s_ig(s_i)$$-$$g'(s_i)\}[E\{G(s_i)\}$$-$$E\{G(z)\}]$$>$0, 
where $g(\cdot)$ is the derivative of $G(\cdot)$, and $g'(\cdot)$ is the derivative of $g(\cdot)$. \mbox{\hfill \qed}
\end{thm}
Note that any independent component $s_i$ satisfying the condition in Theorem \ref{thm:Aapo} is a {\it local} maximum  of $J_G(\bfw)$ but may not correspond to the {\it global} maximum. 
Two conjectures are widely made \cite{Hyva01book}, 
{\bf Conjecture 1}: the condition in Theorem \ref{thm:Aapo} is true for most reasonable choices of $G$ and distributions of the $s_i$;  
{\bf Conjecture 2}: the global maximum of $J_G(\bfw)$ is one of the $s_i$ for most reasonable choices of $G$ and the distributions of the $s_i$. 
In particular, if $G(s)$$=$$s^4$, Conjecture 1 is true for any continuous random variable whose moments exist and kurtosis is non-zero \cite{Hyva99TNN}, and it can also be proven that there are no spurious optima \cite{Delfosse:Loubaton}. Then the global maximum should be one of the $s_i$, {\it i.e.}, Conjecture 2 is true as well. 
However, kurtosis often suffers from sensitivity to outliers. Therefore, more robust functions such as $G(s)$$=$$-\exp (-s^2/2)$ are widely used \cite{Hyva01book}.

\subsection{Linear acyclic causal models}\label{sec:model}
Causal relationships between continuous observed variables $x_i$ ($i$ = 1, $\cdots$, $p$) are typically assumed to be (i) {\it linear} and (ii) {\it acyclic}\cite{Pearl00book,Spirtes93book}. 
For simplicity, we assume that the variables $x_i$ are of zero mean. 
Let $k(i)$ denote such a causal order of $x_i$ that no later variable causes any earlier variable. 
Then, the linear causal relationship can be expressed as
\begin{eqnarray}
x_i := \sum_{k(j)<k(i)} b_{ij} x_j + e_i,\label{eq:model0}
\end{eqnarray}
where $e_i$ is an external influence associated with $x_i$ and is of zero mean. 
(iii) The {\it faithfulness} \cite{Spirtes93book} is typically assumed. 
In this context, the faithfulness implies that correlations and partial correlations of variables $x_i$ are entailed by the graph structure, {\it i.e.}, the zero/non-zero status of $b_{ij}$, not by special parameter values of $b_{ij}$. 
(iv) The external influences $e_i$ are assumed to be independent, which implies there are {\it no unobserved confounders} \cite{Spirtes93book}. 

We emphasize that $x_i$ is equal to $e_i$ if it is not influenced by any other observed variable $x_j$ ($j$$\neq$$i$) inside the model, {\it i.e.}, all the $b_{ij}$ ($j$$\neq$$i$) are zeros. 
That is, an external influence $e_i$ is {\it observed} as $x_i$. 
Then the $x_i$ is called an {\it exogenous observed} variable.\footnote{An exogenous variable is defined as a variable that is not influenced by any other variable inside the model. This definition does not require that it is equal to an external influence. However, in the model (\ref{eq:model0}), any exogenous variable is equal to an external influence due to the acyclicity and no confounder assumptions. }
Otherwise, $e_i$ is called an {\it error}. 
For example, consider the model defined by
\begin{eqnarray*}
x_1 &=& e_1 \\
x_2 &=& 1.5 x_1 + e_2 \\
x_3 &=& 0.8 x_1 -1.3  x_2 + e_3.
\end{eqnarray*}
The $x_1$ is equal to $e_1$ since it is not influenced by either $x_2$ or $x_3$. The $x_1$ is an exogenous observed variable, and $e_2$ and $e_3$ are errors. 
Note that it is obvious that there {\it exists at least one exogenous observed variable} $x_i$($=$$e_i$) due to the acyclicity and no unobserved confounder assumptions. 

In many cases, we do not know anything about the causal relations between the variables $x_i$, and the causal effects $b_{ij}$ are completely unknown. 
Most of previous methods \cite{Pearl00book,Spirtes93book,Shimizu06JMLR} that estimate the entire model structure, {\it i.e.}, all the causal effects $b_{ij}$, 
explicitly or implicitly assume fewer observed variables than observations ($p$$<$$n$). 
Otherwise, the methods are often statistically unreliable. 
In contrast, in the next section, we propose a new method to identify which observed variables are exogenous instead of estimating the entire model structure under the condition $p$$\gg$$n$. 

\section{A new method to identify exogenous variables}\label{sec:method}
In this section, we propose a new method to identify exogenous observed variables. 

\subsection{A new non-Gaussian linear acyclic causal model}\label{sec:ngmodel}
We make two additional assumptions on the distributions of $e_i$ to the model (\ref{eq:model0}) and define a new non-Gaussian linear causal model. 
Let the observed variables $x_i$ in a $p$-dimensional vector be $\bfx$ and external influences $e_i$ in a $p$-dimensional vector $\bfe$.  
Let a $p$$\times$$p$ matrix $\B$ consist of the causal effects $b_{ij}$ where the diagonal elements $b_{ii}$ are all zeros. 
Then the model (\ref{eq:model0}) is written in a matrix form as:
\begin{eqnarray}
\bfx = \B\bfx + \bfe. \label{eq:model}
\end{eqnarray}

Recall that the set of the external influences $e_i$ consist of both exogenous observed variables and errors. 
To distinguish the exogenous variables and errors, we make the following additional assumptions, 
{\bf Assumption 1}: External influences that correspond to exogenous observed variables are non-Gaussian; 
{\bf Assumption 2}: External influences that correspond to errors are non-Gaussian but less non-Gaussian than the exogenous observed variables. 
That is, {\it the model (\ref{eq:model})$=$the model (\ref{eq:model0})$+$Assumptions 1 and 2. }
The first assumption is made to explain why observed data are often considerably non-Gaussian in many  fields \cite{Hyva01book}. 
The second assumption reflects two facts: i) in statistics, errors have been typically considered to arise as sums of a number of unobserved (non-Gaussian) independent  variables, which is why classical methods assume that errors are Gaussian resorting to the central limit theorem; ii) the distinction between Gaussian and non-Gaussian variables is artificial in practice, though. In reality, many variables are not exactly Gaussian. 
Therefore, we allow the errors to be strongly non-Gaussian as long as they are less non-Gaussian than exogenous variables.\footnote{Actually, it is rather easy to show that our algorithm in Section \ref{sec:EggFinder} allows Gaussian errors as well.} 

The distinction between exogenous variables and errors leads to a very simple estimation of exogenous variables proposed in the next subsections.

\subsection{Identification of exogenous variables based on non-Gaussianity and uncorrelatedness}\label{sec:estimation}
We relate the linear non-Gaussian causal model (\ref{eq:model}) with ICA similarly to \cite{Shimizu06JMLR}. 
Let us solve the model (\ref{eq:model}) for $\bfx$ and then we have an ICA model represented by Eq. (\ref{eq:ica}) as follows
\begin{eqnarray}
\bfx &=& (\I-\B)^{-1}\bfe = \A'\bfe. \label{eq:ica2}
\end{eqnarray}
Note that $\I$$-$$\B$ is invertible since it can be permuted to be lower triangular due to the acyclicity assumption if one knew causal orders $k(i)$ \cite{Shimizu06JMLR} and its diagonal elements are all non-zero (unity). 
In the next section we propose a new algorithm to find exogenous variables $x_i$($=$$e_i$) using the relation (\ref{eq:ica2}). 
In this section we present two lemmas that ensures the validity of the algorithm. 

\begin{lem}\label{lemma1}
Assume that the input data $\bfx$ follows the model (\ref{eq:model})\footnote{Namely, we make the following assumptions on the data generating process (i) linearity, (ii) acyclicity, (iii) the faithfulness, (iv) no unobserved confounders (Section \ref{sec:model}), and Assumptions 1 and 2 (Section \ref{sec:ngmodel}). } and that Conjecture 2 (Section \ref{sec:ica}) is true. 
Let us denote by $V_x$ the set of all the observed variables $x_i$. 
Then, the most non-Gaussian observed variable in $V_x$ is exogenous:  
$J(x_i)$ is maximum in $V_x$ $\Rightarrow$ $x_i$$=$$e_i$. \mbox{\hfill \qed}
\end{lem} 

\paragraph{Proof}
Eq. (\ref{eq:ica2}) shows that the model (\ref{eq:model}) is an ICA model, where external influences $e_i$ are independent components (ICs). 
The set of the external influences consist of exogenous observed variables and errors.  
Due to the model assumption (Assumption 2 in Section \ref{sec:ngmodel}), exogenous observed variables are more non-Gaussian than errors. 
Therefore, the most non-Gaussian {\it exogenous} observed variable is the most non-Gaussian IC.  
Next, according to Conjecture 2 that is here assumed to be true, the most non-Gaussian IC, {\it i.e.}, the most non-Gaussian {\it exogenous} observed variable, is the global maximum of the non-Gaussianity measure $J(\bfw^T\bfx)$$=$$J_G(\bfw)$ among such linear combinations of observed variables $\bfw^T\bfx$ with the constraint $E\{(\bfw^T\bfx)^2\}$$=$1, which include all the observed variables $x_i$ in $V_x$. 
Therefore, the most non-Gaussian observed variable is the most non-Gaussian {\it exogenous} variable.  \mbox{\hfill \bsquare}


\begin{lem}\label{lemma2}
Assume the assumptions of Lemma \ref{lemma1}. 
Let us denote by $E$ a strict subset of exogenous observed variables so that it does not contain at least one exogenous variable. 
Let us denote by $U_E$ the set of observed variables uncorrelated with any variable in $E$. 
Then the most non-Gaussian observed variable in $U_E$ is exogenous: 
$J(x_i)$ is maximum in $U_E$ $\Rightarrow$ $x_i$$=$$e_i$. \mbox{\hfill \qed}
\end{lem}

\paragraph{Proof}
First, the set $V_x$ is the union of three disjoint sets: $E$, $U_E$ and $C_E$, where $C_E$ is the set of observed variables in $V_x\backslash E$ correlated with a variable in $E$. 
By definition, any variable in $U_E$ are not correlated with any variable in $E$. 
 Since the faithfulness is assumed, the zero correlations are only due to the graph structure. 
Therefore, there is no directed path from any variable in $E$ to any variable in $U_E$. 
Similarly, there is a directed path from each (exogenous) variable in $E$ to a variable in $C_E$. 
 Next, there can be no directed path from any variable in $C_E$ to any variable in $U_E$. 
Otherwise, there would be a directed path from such a variable in $E$ from which there is a directed path to a variable in $C_E$ to a variable in $U_E$ through the variable in $C_E$. 
Then, due to the faithfulness, the variable in $E$ must correlate with the variable in $U_E$, which contradicts the definition of $U_E$.

To sum up, there is no directed path from any variable in $E \cup C_E$ to any variable in $U_E$.  
Since any directed path from the external influence $e_i$ associated with any variable $x_i$ in $V_x$ must go through the $x_i$, 
there is no directed path from the external influence associated with any variable in $E \cup C_E$ to any variable in $U_E$. 
In other words, there can be directed paths from {\it only} the external influences associated with any variables in $U_E$ to some variables in $U_E$. 
Then, we again have an ICA model: $\widetilde{\bfx}$$=$$\widetilde{\A}'$$\widetilde{\bfe}$, where 
 $\widetilde{\bfx}$ and $\widetilde{\bfe}$ are vectors whose elements are the variables in $U_E$ and corresponding external influences in $\bfe$ in Eq. (\ref{eq:ica2}), and 
$\widetilde{\A}'$ is the corresponding submatrix of $\A'$ in Eq. (\ref{eq:ica2}). 
Recursively applying Lemma \ref{lemma1} shows that the most non-Gaussian variable in $U_E$ is exogenous. \mbox{\hfill \bsquare}


To find uncorrelated variables, we simply use the ordinary Gaussianity-based testing method \cite{Lehmann05Test} and control the false discovery rate \cite{Benjamini95JRSSB} to 5\% for multiplicity of tests. 
Though non-parametric methods \cite{Lehmann05Test} is desirable for more rigorous testing in the non-Gaussian setting, we used the Gaussian method that is more computationally efficient and seems to work relatively well in our simulations. Future work would address what is the better testing procedure taking non-Gaussianity into account. 

\subsection{Exogenous generating variable finder: EggFinder}\label{sec:EggFinder}
Based on the discussions in the previous subsection, we propose an algorithm to find exogenous variables one by one, which we call EggFinder (ExoGenous Generating variable Finder):

\noindent
	 \begin{enumerate}
	 \item Given $V_x$, initialize $E$$=$$\emptyset$, $U_E^{(1)}$$=$$V_x$, and $m$$:=$1.  
	\item \label{step:1} Repeat until no variables $x_i$ are uncorrelated with exogenous variable candidates, {\it i.e.}, $U_E^{(m)}$$=$$\emptyset$:
	\begin{enumerate}
	\item Find the most non-Gaussian variable $x_m$ in $U_E^{(m)}$: 
	\begin{eqnarray}
	x_m=\arg \max_{x \in U_E^{(m)}} J(x), 
	\end{eqnarray}
	 where $J$ is the non-Gaussianity measure in Eq. (\ref{eq:approxJ}) with 
	 \begin{eqnarray}
	 G(x)=-\exp(-x^2/2).
	 \end{eqnarray} 	
	\item Add the most non-Gaussian variable $x_m$ to $E$, that is, $E$$=$$E$$\cup$$\{x_m\}$. 
	\item \label{step:uncorr} Let  $U_E^{(m+1)}$ to be the set of variables $x_i$ uncorrelated with any variable in $E$, and $m$$:=$$m$$+$$1$. 
	\end{enumerate}
	 \end{enumerate}

In Step \ref{step:uncorr}, we use the Gaussianity-based testing method and control the false discovery rate to 5\%.

\section{Experiments on artificial data}\label{sec:exp}
We performed two experiments with artificial data 
to evaluate the performance of EggFinder when $p$$\gg$$n$ (Experiment 1) and its scalability (Experiment 2). 
The experiments were conducted on a PC equipped with two 2.8 GHz Quad-Core Intel Xeon processors and 2GB memory using Matlab 7.6.

\subsection{Experiment 1: Performance evaluation when p$\gg$n}\label{sec:exp-reliability}
We studied the performance of EggFinder when $p$$\gg$$n$ under a linear non-Gaussian acyclic model having a sparse graph structure and various degrees of error non-Gaussianity. 
Many real-world networks such as gene networks are often considered to have scale-free graph structures. However, as far as we know, there is no standard way to create a {\it directed} scale-free graph. Therefore, we first randomly created a (conventional) sparse directed acyclic graph with $p$$=$1,000 variables using a standard software Tetrad (http://www.phil.cmu.edu/projects/tetrad/). 
The resulting graph contained 1,000 edges and $\ell$$=$171 exogenous variables. 
We randomly determined each element of the matrix $\B$ in the model (\ref{eq:model}) to follow this graph structure and make the standard deviations of $x_i$ owing to parent observed variables ranged in the interval $[0.5,1.5]$. 

We generated non-Gaussian exogenous variables and errors as follows. 
We randomly generated a non-Gaussian exogenous observed variable $x_i$($=$$e_i$) that was sub- or super-Gaussian with probability 50\%.\footnote{We first generated a Gaussian variable $z_i$ with zero mean and unit variance and subsequently transformed it to a non-Gaussian variable by $s_i={\rm sign}(z_i)|z_i|^{q_i}$. 
The nonlinear exponent $q_i$ was randomly selected to lie in $[0.5, 0.8]$ or $[1.2, 2.0]$ with probability 50\%. 
The former gave a sub-Gaussian symmetric variable, and the latter a super-Gaussian symmetric variable.
Finally, the transformed variable $s_i$ was scaled to the standard deviation randomly selected in the interval $[0.5,1.5]$ and was taken as an exogenous variable. } 
Next, for each error $e_i$, we randomly generated $h$ ($h$$=$1, 3, 5 and 50) non-Gaussian variables having unit variance in the same manner as for exogenous variables and subsequently took the sum of them. 
We then scaled the sum to the standard deviation selected similarly to the cases of exogenous variables and finally took it as an error $e_i$. 
A larger $h$ (the number of non-Gaussian variables summed) would generate a less non-Gaussian error due to the central limit theorem. 

\begin{figure*}
\begin{center}
\includegraphics[width=4.75in]{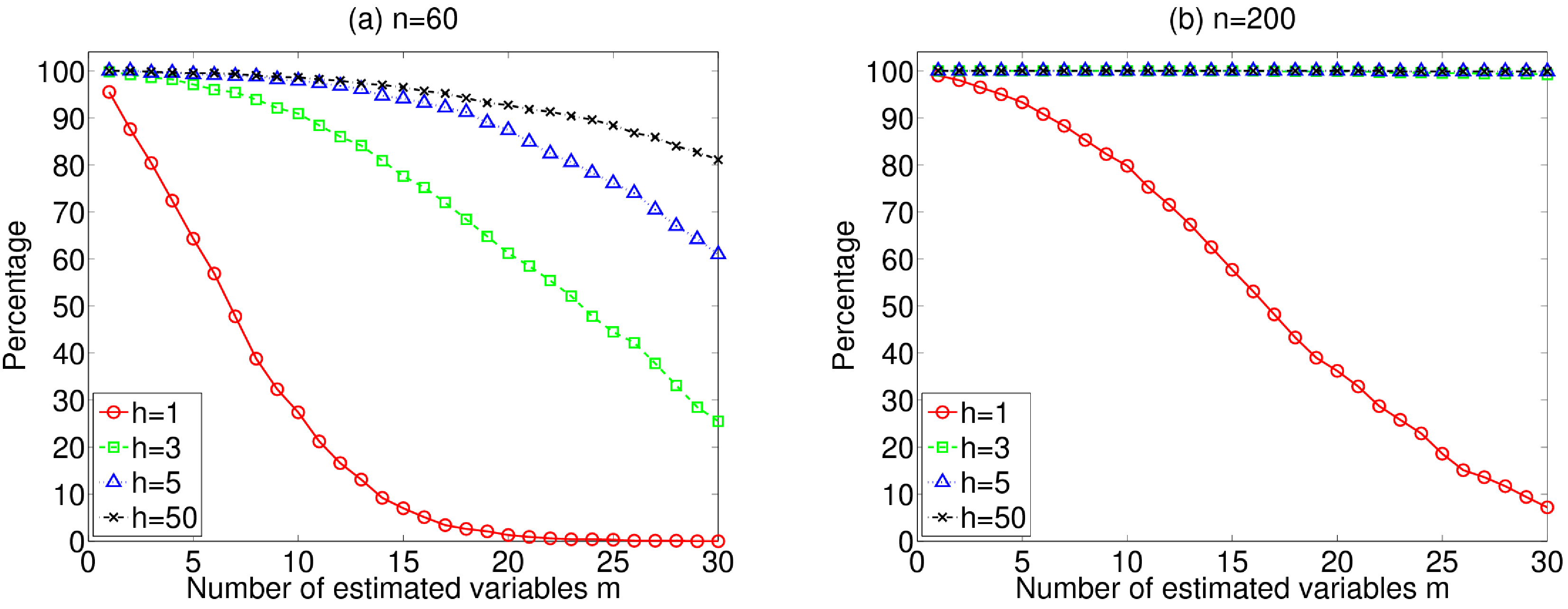}
\end{center}
\caption{Percentages of datasets where all the top $m$ estimated variables were actually exogenous under (a) $n$$=$60; (b) $n$$=$200.}
\label{fig:percentage}
\end{figure*}

Finally, we randomly generated 1,000 datasets under each combination of $h$ and $n$ ($n$$=$30, 60, 100 and 200) and fed the datasets to EggFinder. 
For each combination, we computed percentages of datasets where all the top $m$ estimated variables were actually exogenous. 
In Figure~\ref{fig:percentage}, the relations between the percentage and $m$ are plotted for some representative conditions due to the limited space. 
First, in all the conditions the percentages monotonically decrease when $m$ increases.  
Second, the percentages generally increase when the sample size $n$ increases.  
Similar changes of the percentages are observed when the errors are less non-Gaussian. 
This is reasonable since a larger $n$ enables more accurate estimation of non-Gaussianity and correlation, and a larger $h$ generates data more consistent with the assumptions of the model (\ref{eq:model}). 
In summary, EggFinder successfully finds a set of exogenous variables up to more than $m$$=$10 in many practical conditions. 
However, EggFinder may not find all the exogenous variables when $p$$\gg$$n$, although it asymptotically finds all the exogenous variables if all the assumptions made in Lemmas \ref{lemma1} and \ref{lemma2} hold. 

Interestingly, EggFinder did not fail completely and identified a couple of exogenous variables even for the $h$$=$1 condition where the distributional assumption on errors was most likely to be violated. 
This is presumably because the endogenous variables are sums of non-Gaussian errors and exogenous variables, so due to the central limit theorem they are likely to be less non-Gaussian than the exogenous variables, even if the errors and exogenous variables have the same degree of non-Gaussianity. 

\subsection{Experiment 2: A scalability comparison}\label{sec:exp-scalability}
To evaluate the scalability of EggFinder, we made a comparison with LiNGAM \cite{Shimizu06JMLR} that also utilizes non-Gaussianity but estimates the entire model structure. 
LiNGAM is applicable to the conditions where the errors are more non-Gaussian than exogenous variables unlike EggFinder. 
Once LiNGAM finds causal orders $k(i)$ of variables, edges having some possibility of zero effects are pruned using the bootstrapping method \cite{Efron93book}. 
Exogenous variables are determined as origins of the causal chains. 

We randomly created three directed acyclic graphs with $p$$=$50, 100 and 150 variables similarly to Experiment 1. The resulting numbers of exogenous variables $\ell$ were 14, 30 and 41 for the three graphs, respectively. 
For each graph, we generated a matrix $\B$ and selected non-Gaussian distributions of external influences in the same manner as Experiment 1. 
A rather strong degree of error non-Gaussianity under $h$$=$3 was tested to make the comparison as fair as possible. 
Then we randomly generated 50 datasets with the sample size 500 under each $p$ and fed the datasets to EggFinder and LiNGAM. 
In LiNGAM, the resampling size for the bootstrapping was 5,000. 
 
The medians of elapsed time, precision and recall are shown in Table \ref{Tab:exp-scalability}. 
Precision is the proportion of correctly found exogenous variables to estimated exogenous variable candidates. Recall is the proportion of correctly found exogenous variables to all the actual exogenous variables. 
We see that EggFinder was much faster than LiNGAM for all the conditions and scored much better precision and recall than LiNGAM. The poor performance of LiNGAM is easily understood since it had to estimate much more parameters (already 2,500 for $p$$=$50) than the sample size available ($n$$=$500). 

\begin{table}[time]
\caption{Medians of elapsed time, precision and recall}
\label{Tab:exp-scalability}
\begin{center}
\begin{tabular}{cc||ccc|ccc}
 & & \multicolumn{3}{c|}{\bf EggFinder} & \multicolumn{3}{c}{\bf LiNGAM} \\
$p$ & $\ell$ & \hspace{3pt}Time & Prec. & Rec. & Time & Prec. & Rec. \\
\hline 
50 & 14 & 1.35  & 0.80 & 0.40 & \hspace{5pt} 297.47 (41.16) & 0.50 & 0.10 \\
100 & 30 & 1.92  & 0.84 & 0.95 & \hspace{5pt} 876.05 (29.67) &  0.32 & 0.10 \\
150 & 41 & 2.20 & 0.88 & 0.90  & 2,104.90 (76.46) & 0.30 & 0.07 \\
\multicolumn{8}{c}{} \\
\multicolumn{8}{l}{Median elapsed times of LiNGAM with (without) bootstrapping are shown. }
\end{tabular}
\end{center}
\end{table}

\section{Application to microarray gene expression data}\label{sec:genedata}
To evaluate the practicability of EggFinder, we analyzed a real-world dataset of gene microarray collected in an experiment on 60 mice \cite{Kanno06BMCG}. 
The aim of the experiment was to dose a chemical compound called 2,3,7,8-tetrachlorodibenzo-$p$-dioxin (TCDD) to mice and to see how it affects the gene expression levels of the liver cells. 
The experiment was conducted with completely random sampling of mice under every combination of two factors. 
The first factor was the concentration of TCDD (0,1,3,10 and 30 microg/kg), and the second factor was the elapsed time since it was dosed (2, 4, 8 and 24 hours). The total number of experimental conditions was 20. 
For each condition, gene expression levels of 45,101 genes in the liver cells of three mice were measured using GeneChip microarrays. The total number of observations was 60. We used the data normalized using the `Per cell' method proposed by the authors of \cite{Kanno06BMCG} (http://www.biomedcentral.com/1471-2164/7/64).

As a standard preprocessing, we first conducted $t$-tests for the differences of means \cite{Lehmann05Test} of the gene expression levels between the lowest and second lowest concentration conditions of TCDD at 2 hours elapsed time. We then selected 1,000 genes that expressed the most significance since such genes were likely to relevant to the TCDD dosing. 
Next, as mentioned in the previous paragraph, the dataset consisted of 60 observations randomly sampled from 20 different populations (three mice for each condition). 
However, the sample size three was too small to analyze each population separately.
Therefore, we made a relatively common assumption in Bayesian networks \cite{Muthen89PMK} that the populations might have different mean structures but had the same dependency structures. 
Then it is expected to be fairly well validated to combine all the data into a single dataset and analyze the dependency structures after computing the means for each population and subtracting them. 
Thus, we obtained a data matrix with the number of variables $p$$=$1,000 and the sample size $n$$=$60. 


Then we applied EggFinder to the combined data and found 42 exogenous variable candidates. 
We further computed bootstrap probabilities \cite{Felsenstein85Evol} to assess the statistical reliability. 
Here the bootstrap probability for a gene was defined as the proportion of how many times the gene belonged to estimated sets of exogenous variables to the resampling size 5,000. 
Since the sample size for each population was quite small (three), the estimated bootstrapping probabilities might not be very accurate. However, we thought that it was more useful to examine them than not. 
Then 26 exogenous variable candidates (trigger genes) were significantly often found with the significance level 5\%. 

First, we mention that it is hardly known which genes TCDD first affects and how it triggers the gene network. 
This is why we applied EggFinder to find promising trigger gene candidates of novel TCDD-induced pathways in the gene network. 
Fortunately, some background knowledge in biology is available to assess the performance of EggFinder. 
In mouse cells, TCDD binds to a AhR (Aryl hydrocarbon Receptor). 
TCDD-bound AhR works as a transcription factor and binds a specific short sequence called XRE (Xenobiotic Responsive-Element). 
Thus, a possible way to find such genes that TCDD first affects is to see if there are XREs in the DNA sequences of genes. 
However, since XRE is a short sequence, it can be found in various positions in genome, even in the DNA sequences of genes that TCDD does not first affect. 
Therefore, simple sequence analysis based on sequence pattern matching
tends to produce a number of false positive genes that have irrelevant XREs for TCDD. 
Further, TCDD could first affect genes having no XREs as well.
Since EggFinder works well in simulations (Section~\ref{sec:exp}), 
it is expected that EggFinder finds better trigger gene candidates having relevant XREs or possible other sequences for TCDD if the model assumptions in Eq. (\ref{eq:model}) are fairly reasonable. 

EggFinder found seven genes having XREs in their promoters (until 1,000 base pairs upstream from the transcription start sites) or their exons according to a database called DRGdb-NIES.\footnote{http://www.nies.go.jp/health/drgdb/drgdb-top/TOP.htm} 
Two of the seven genes, \verb+1422217_a_at+ (CYP1A1) and  \verb+1428288_at+, are known or conjectured in biology to be causally related to TCDD. 
First, CYP1A1 is a well-known dioxin response gene and assigned as dibenzo-$p$-dioxin metabolic process in a database called Gene Ontology (http://www.geneontology.org/index.shtml). 
Second, \verb+1428288_at+ is related to a progesterone receptor signaling pathway according to the Gene Ontology database. Such a progesterone receptor of rats is known to be inhibited by TCDD. 
Although there is not enough background knowledge about the other five genes yet, 
the seven genes with XREs could be good candidates to intervene on when conducting experiments to find trigger genes of TCDD-induced pathways in the gene network. 

 
Many real-world causal networks including gene networks are known to be nonlinear and cyclic \cite{Bernardo05Nature}. An important question for future research is to investigate how seriously this harms the performance of EggFinder that assumes linearity and acyclicity and, eventually, how it can be solved or alleviated.



\section{Conclusion}\label{sec:conc}
We proposed a method to find exogenous variables in a linear non-Gaussian causal model in data with orders of magnitude more variables than observations. 
In the simulations the new method successfully found more exogenous variables and was more computationally efficient than a previous method. 
This would be an important first step for developing advanced causal analysis methods in the challenging situations $p$$\gg$$n$.
Important topics for future research are i) how the number of valid exogenous variable candidates is estimated well; 
ii) how we could evaluate model fit; 
iii) how the model assumptions can be relaxed towards more general nonlinear modeling. 

\bibliography{Shohei_Tex_Ref}
\bibliographystyle{splncs}

\end{document}